\documentclass[10pt,journal,compsoc]{IEEEtran}

\usepackage[T1]{fontenc}
\usepackage{amsmath,amsfonts}
\usepackage{algorithmic}
\usepackage{algorithm}
\usepackage{array}
\usepackage[caption=false,font=normalsize,labelfont=sf,textfont=sf]{subfig}
\usepackage{textcomp}
\usepackage{stfloats}
\usepackage{placeins}
\usepackage{url}
\usepackage{verbatim}
\usepackage{graphicx}
\usepackage[nocompress]{cite}
\usepackage{booktabs}
\usepackage{tabularx}
\usepackage[hidelinks]{hyperref}

\newcommand{\pcdBest}[1]{\ensuremath{\mathbf{#1}}}
\newcommand{\pcdSecond}[1]{\ensuremath{\underline{#1}}}
\newcommand{\pcdThird}[1]{\ensuremath{\mathit{#1}}}

\begin{document}
\bstctlcite{pcd-bib-control}

\title{PCD: Predictive Coding Dynamics for Irregularly Sampled Time Series}


\author{Tianrui~Gou, Jiayangmei~Xu, Xingzhou~Zhang, and~Lei~Sha%
\thanks{This work has been submitted to the IEEE for possible publication. Copyright may be transferred without notice, after which this version may no longer be accessible. \par\medskip This work was supported by the National Science Fund
for Excellent Young Scholars (Overseas) under Grant KZ37117501;
the National Natural Science Foundation of China under Grant 62306024;
the National Cyber Security-National Science and Technology Major
Project under Grant 2025ZD1503602;
the New Generation Artificial Intelligence-National Science and
Technology Major Project under Grants 2026ZD0128000 and 2026ZD0128003;
the Fundamental Research Funds for the Central Universities;
and the Beijing Advanced Innovation Center for Future Blockchain
and Privacy Computing.
(Corresponding author: Lei Sha.)}%
\thanks{Tianrui Gou is with the School of Artificial Intelligence,
Beihang University, Beijing 100191, China.}%
\thanks{Jiayangmei Xu is with the Department of Foreign Languages,
Beihang University, Beijing 100191, China.}%
\thanks{Xingzhou Zhang is with the School of Artificial Intelligence,
Beihang University, Beijing 100191, China.}%
\thanks{Lei Sha is with the School of Artificial Intelligence,
Beihang University, Beijing 100191, China
(e-mail: shalei@buaa.edu.cn).}%
}

\markboth{PCD: Predictive Coding Dynamics for Irregularly Sampled Time Series}%
{Gou \MakeLowercase{\textit{et al.}}: Predictive Coding Dynamics}






\IEEEtitleabstractindextext{%
    \begin{abstract}
        Dynamic systems evolve continuously over time, whereas observations in real scenarios are often sparse, irregular, and asynchronous. Accurate prediction therefore requires inferring latent state evolution between observations, as future values depend on partially observed trajectories. We propose Predictive Coding Dynamics (PCD), a residual feedback mechanism for irregular continuous dynamic models. PCD maintains previous reconstructions, predicts observed entries from them, and incorporates masked signed residuals into state recognition through an energy penalty. The mechanism is implemented with ODE-RNN and Latent ODE, supporting event-based hidden updates and latent state recognition. Experiments on forecasting and trajectory recovery tasks show that PCD improves continuous dynamic backbones under sparse and irregular sampling. The results further indicate that predictive error feedback improves latent state estimation when observations constrain the current trajectory. These results suggest that PCD provides a residual feedback framework that jointly supports trajectory generation, observation consistency checking, and state refinement in continuous-time series modeling. 
        The source code is available at
        \url{https://anonymous.4open.science/r/PCD-for-paper-92F1/}.
    \end{abstract}

}

\maketitle
\IEEEdisplaynontitleabstractindextext

\IEEEraisesectionheading{%
    \section{Introduction}\label{sec:introduction}
}

\IEEEPARstart{D}{ynamic} systems evolve continuously, while real-world measurements are typically collected at finite and irregular time points.
Such irregular observations arise in clinical monitoring, physical simulation, weather sensing, industrial control, and human activity analysis.
Between observations, the system evolves through unobserved states, and models must infer the corresponding latent transitions.
Pointwise prediction ignores this evolution and provides incomplete representations of latent states.

Irregular sampling complicates inference because elapsed time provides constraints but is insufficient to determine state transitions.
Short intervals often imply local smoothness, while long intervals may contain nonlinear changes, abrupt variations, or multiple hidden phases.
In multivariate series, variables are often observed asynchronously, requiring models to infer dependencies from incomplete temporal evidence.
Values, timestamps, masks, and sampling density jointly provide evidence for latent trajectory reconstruction.

Existing irregular time-series methods mainly address missing observations and continuous latent dynamics.
GRU-D introduces temporal decay to capture missing measurement effects on recurrent states~\cite{pcd01}.
Li and Marlin analyze irregular sampling from the perspective of missing data and show that timestamps and observation masks provide information beyond observed values~\cite{pcd02}.
These studies highlight the role of explicitly modeling observation patterns.

Continuous-time neural models directly describe latent trajectory evolution.
Neural ODE models latent dynamics using differential equations and estimates states at arbitrary query times~\cite{pcd03}.
Latent ODE extends this approach to irregular sequences by learning latent initial states and continuous trajectories~\cite{pcd04}.
GRU-ODE-Bayes applies observation-driven Bayesian updates for sparse observations~\cite{pcd05}, whereas Neural CDE represents irregular observations as continuous control signals~\cite{pcd06}.
Recent studies explore continuous-time attention, reconstruction-based representation learning, and recurrent continuous units for irregular prediction~\cite{pcd07,pcd08,pcd09}.

Despite architectural differences, most methods follow a similar inference paradigm.
Observations are encoded into latent states or distributions, propagated through continuous dynamics, and subsequently decoded into trajectories.
Training minimizes reconstruction discrepancies between trajectories and observations.
However, prediction errors are typically used as optimization signals rather than feedback for state estimation.

Prediction residuals can therefore be exploited as additional inference signals.
A reconstructed trajectory provides a temporary estimate of system evolution.
When it disagrees with observations, signed residuals indicate both correction magnitude and direction.
Such feedback is valuable under sparse sampling, where multiple latent trajectories may explain the observations while leading to different future predictions.

Predictive coding provides a theoretical basis for residual-based refinement.
Rao and Ballard introduced predictive coding as a hierarchical framework in which prediction errors update internal representations~\cite{pcd10}.
Friston connected prediction errors with hierarchical generative inference and free energy minimization, in which internal states are revised according to prediction-observation discrepancies~\cite{pcd11}.
Thus, prediction error provides a mechanism for refining latent representations beyond measuring model discrepancy.

Recent studies explore predictive coding as trainable neural architectures.
PredNet uses recurrent prediction-error propagation for video prediction and representation learning~\cite{pcd12}.
Deep predictive coding networks combine top-down predictions and bottom-up errors for object recognition~\cite{pcd13}, while local recurrent predictive coding explicitly models prediction and error units~\cite{pcd14}.
Predictive coding has also been linked to optimization and learning dynamics.
Song et al.\ show that predictive coding networks can implement backpropagation through activity inference~\cite{pcd15}, and Salvatori et al.\ extend this connection to reverse differentiation over computation graphs~\cite{pcd16}.

Predictive coding has been studied in sequential and structured learning.
Temporal predictive coding improves sequential memory through recurrent prediction and error correction~\cite{pcd17}.
Incremental predictive coding develops stable learning procedures~\cite{pcd18}, while divide-and-conquer predictive coding formulates predictive coding as structured Bayesian inference~\cite{pcd19}.
Recent predictive feedback methods apply future-oriented discrepancies to time-series forecasting~\cite{pcd20}.
These studies demonstrate that prediction errors can facilitate representation refinement, temporal learning, and adaptive inference.

However, existing predictive coding methods mainly target representation learning, biological modeling, or task-specific prediction.
They do not directly refine latent continuous trajectories under irregular observations.
For irregular dynamics, reconstructed trajectories can be evaluated at observed timestamps.
Residuals between reconstructed states and observations provide sample-specific signals for continuous state recognition.

We propose Predictive Coding Dynamics (PCD), a residual feedback mechanism for irregular continuous dynamic models.
PCD first reconstructs trajectories using a continuous dynamic backbone, stores previous reconstructions in a cache, and forms internal predictions at observed times.
It computes signed residuals between predictions and observations, removes missing entries with observation masks, and applies variable weights to control feedback across dimensions.
The residual channel is incorporated into the recognition process, enabling latent state inference from both observations and reconstruction errors.

We implement PCD with ODE-RNN and Latent ODE.
ODE-RNN propagates hidden states between observation events and updates them when events arrive; PCD injects residuals into this update.
Latent ODE infers an initial latent distribution from observations and generates trajectories with an ODE solver; PCD injects residuals into its recognition network.
Both implementations follow the same refinement procedure: reconstruction, cached prediction, masked signed residual computation, state recognition, and reconstruction update.

This work makes three contributions.
\begin{enumerate}
    \item We formulate irregular continuous-time modeling as an iterative error-feedback inference process, in which discrepancies between internally reconstructed trajectories and asynchronous observations are explicitly reused to revise latent states, rather than serving only as optimization losses.

    \item We develop a backbone-agnostic predictive coding mechanism that constructs masked signed residuals from cached trajectories and injects them into continuous-time state recognition.
    We instantiate this mechanism in both event-driven ODE-RNN and latent-variable Latent ODE, demonstrating its applicability to distinct continuous-time inference paradigms.

    \item We evaluate PCD on forecasting and trajectory recovery tasks across simulated, clinical, motion, meteorological, and regular forecasting benchmarks, showing that error feedback provides additional benefits when observations impose strong constraints on latent dynamics.
\end{enumerate}

The rest of this paper is organized as follows.
Section~\ref{sec:related-work} reviews irregular time-series models and predictive coding methods.
Section~\ref{sec:methodology} presents PCD, including residual construction, backbone integration, optimization, and cache update.
Section~\ref{sec:experiments} reports experiments and ablation analysis.
Section~\ref{sec:conclusion} concludes and discusses limitations.

\section{Related Works}
\label{sec:related-work}

\subsection{Predictive Coding and Predictive-Feedback Learning}
\label{subsec:predictive-coding}

Predictive coding models refine internal representations by reducing discrepancies between predictions and observations.
Rao and Ballard proposed a hierarchical framework in which prediction errors update internal states across representation levels~\cite{pcd10}.
Friston formulated prediction errors within hierarchical generative inference and related predictive coding to free-energy minimization and adaptive state estimation~\cite{pcd11}.
These studies indicate that prediction errors can provide inference signals beyond conventional evaluation metrics.

Recent work incorporates predictive coding into trainable neural architectures.
PredNet employs recurrent prediction and error units for video prediction, where hierarchical errors are used to update internal representations~\cite{pcd12}.
Deep predictive coding networks apply top-down predictions and bottom-up error propagation to object recognition tasks~\cite{pcd13}.
Local recurrent predictive coding networks explicitly model interactions between prediction and error units~\cite{pcd14}.
Predictive coding has also been studied from perspectives of biological plausibility and optimization.
Lotter et al.\ reported that prediction-trained recurrent networks reproduce properties of biological neural responses~\cite{pcd21}, while Song et al.\ showed that predictive coding dynamics can approximate backpropagation through iterative activity inference~\cite{pcd15}.

Recent works study predictive coding as a learning and inference framework.
Salvatori et al.\ analyze predictive coding through reverse differentiation over computation graphs~\cite{pcd16}.
Temporal predictive coding applies recurrent error correction to sequential memory~\cite{pcd17}, while goal-conditioned predictive coding models trajectory representations for offline reinforcement learning~\cite{pcd22}.
Incremental predictive coding investigates optimization stability~\cite{pcd18}, and divide-and-conquer predictive coding formulates predictive coding within a structured Bayesian inference framework~\cite{pcd19}.
Other studies examine optimization behavior and predictive feedback mechanisms~\cite{pcd23,pcd24,pcd20}.
These methods focus on representation learning, optimization, and specific prediction tasks.
They provide limited mechanisms for refining continuous latent trajectories under irregular observations, where prediction errors can provide additional signals for state recognition.

\subsection{Continuous-Time Modeling for Irregular Observations}
\label{subsec:continuous-time}

Continuous-time neural models represent systems whose dynamics evolve between observations.
Neural ODE introduces differential equations into neural networks and enables latent state evaluation at arbitrary timestamps~\cite{pcd03}.
Latent ODE applies this formulation to irregular sequences by learning latent initial states and generating continuous trajectories~\cite{pcd04}.
ODE-RNN combines recurrent observation updates with continuous latent propagation for irregular events~\cite{pcd04}.
GRU-ODE-Bayes incorporates probabilistic state updates upon new observations to model uncertainty in sparse measurements~\cite{pcd05}.

Several studies enhance continuous-time models by introducing additional temporal representations and dynamic mechanisms.
Neural CDE models irregular observations as continuous control paths and infers hidden dynamics from the resulting trajectories~\cite{pcd06}.
Neural SDE introduces stochastic differential equations to model uncertainty in partially observed systems~\cite{pcd25}.
Continuous recurrent units separate irregular time evolution from recurrent updates~\cite{pcd09}.
Neural flows model latent transformations through continuous flows and improve latent state propagation~\cite{pcd26}.
ContiFormer integrates Transformer architectures with continuous-time attention to model interactions among irregular observations~\cite{pcd07}.

Recent methods introduce additional structures for irregular forecasting.
Methods based on trajectory reconstruction use prediction errors to improve latent representations from sparse observations~\cite{pcd08}.
Variants of controlled differential equations improve trajectory representation and extrapolation in irregular forecasting and classification tasks~\cite{pcd27,pcd28}.
Stable stochastic differential equation models analyze continuous stochastic dynamics under irregular sampling~\cite{pcd29}.
These methods represent irregular trajectories through continuous dynamics.
Their reconstruction errors are mainly used during optimization and provide limited information for latent state updates.

\subsection{Observation Representation for Irregular Time Series}
\label{subsec:observation-representation}

Representing incomplete and asynchronous observations remains an important problem in irregular time-series modeling.
Early methods handle missing values through recurrent state correction.
GRU-D introduces temporal decay factors to model the effects of missing intervals on hidden states~\cite{pcd01}.
BRITS formulates missing-value reconstruction through bidirectional recurrent imputation and learns latent representations during this process~\cite{pcd30}.
Li and Marlin analyze irregular sampling as a missing-data problem and demonstrate that timestamps and observation masks provide additional information for model learning~\cite{pcd02}.

Later methods develop representations designed for irregular observations.
SeFT treats irregular measurements as unordered sets and learns permutation-invariant representations~\cite{pcd31}.
Multi-time attention networks aggregate asynchronous measurements at arbitrary query times using attention~\cite{pcd32}.
Raindrop uses graph structures to model relationships among variables and observation patterns in irregular clinical data~\cite{pcd33}.
These studies demonstrate that observation structures provide additional information for modeling irregular time series.

Recent methods introduce structural models for multivariate irregular forecasting.
Graph methods model dependencies among variables through relational structures.
t-PatchGNN organizes irregular series into local patches and graphs to capture temporal and variable interactions~\cite{pcd34}.
HyperIMTS models higher-order variable relations with hypergraphs~\cite{pcd35}.
Hi-Patch introduces hierarchical patch representations for irregular multivariate sequences~\cite{pcd36}.
Other studies investigate models that capture global temporal dependencies.
Transformer models investigate continuous attention mechanisms and patch representations for irregular prediction~\cite{pcd37,pcd38,pcd39}.
Graph neural network methods model multivariate dependencies through relational structures and spectral information~\cite{pcd40,pcd41,pcd42,pcd43,pcd44}.
Recent pretrained models study the use of large language models for transferring sequence information to irregular forecasting tasks~\cite{pcd45,pcd46,pcd47}.
These methods improve the encoding of observations and structural information.
However, their inference process primarily relies on observed inputs, and the differences between generated trajectories and observations are rarely used for iterative state refinement.

\section{Methodology}
\label{sec:methodology}

\begin{figure*}[!tp]
    \centering
    \includegraphics[width=0.95\textwidth]
        {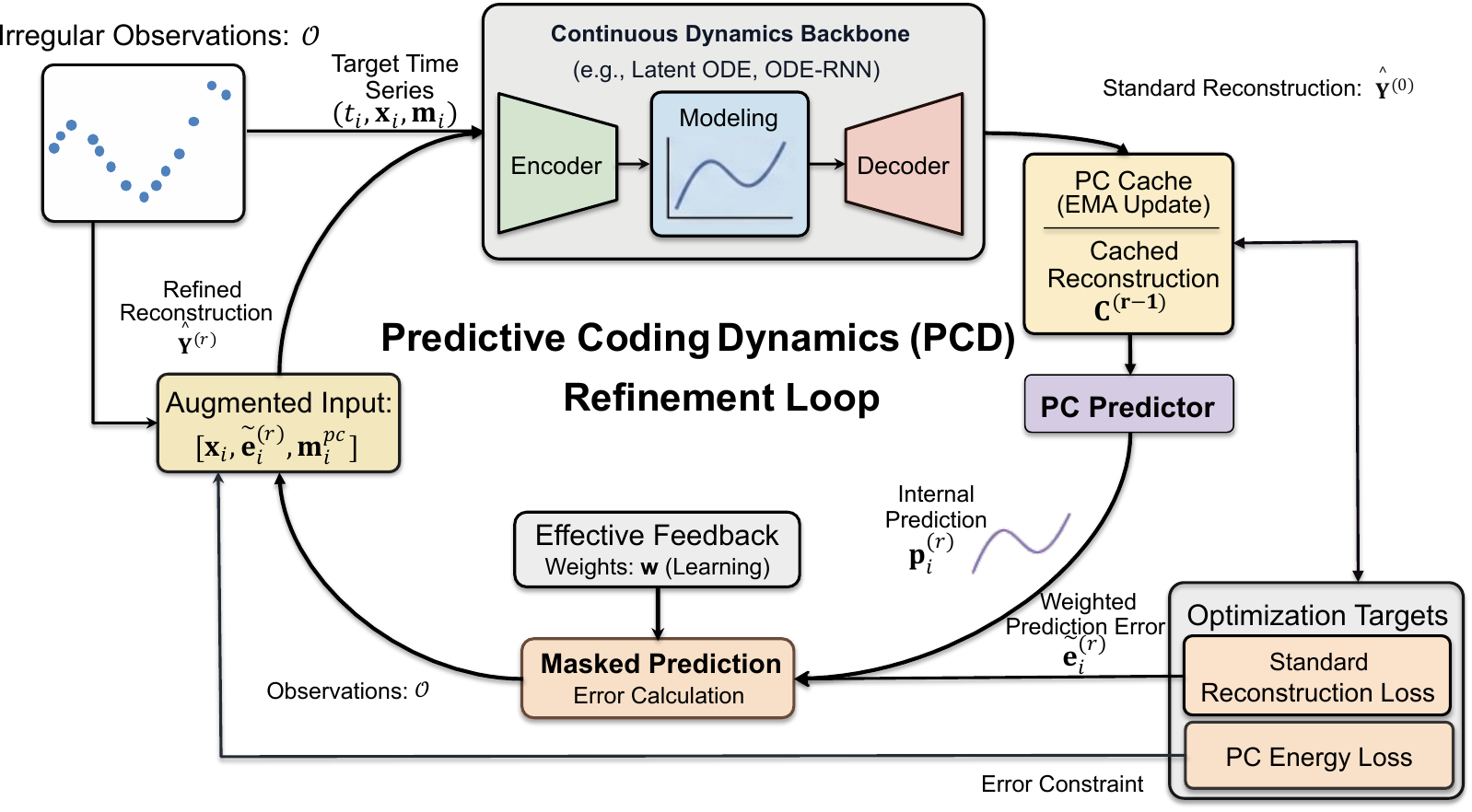}
    \caption{Predictive Coding Dynamics (PCD) feedback loop
    for irregular continuous-time modeling.
    The backbone reconstructs a trajectory, the PC predictor
    generates predictions from the cached reconstruction,
    and weighted masked residuals are incorporated into
    state recognition for the next reconstruction.}
    \label{fig:pcd-framework}
\end{figure*}

As shown in Fig.~\ref{fig:pcd-framework}, the continuous-time backbone generates a reconstruction trajectory through recognition, latent-state propagation, and decoding.
PCD stores previous reconstructions in a cache using an exponential moving average.
The PC predictor uses the cached trajectory to generate predictions at observed timestamps.
The difference between internal predictions and observations is masked to retain errors only at observed dimensions.
The masked error is used as feedback during subsequent state estimation.

During the $r$-th refinement step, learnable feedback weights scale the masked prediction error for each variable, and the scaled error is concatenated with the original observations.
The resulting input is provided to the same continuous-time backbone, allowing prediction errors to influence state recognition during trajectory reconstruction.
In ODE-RNN, the error signal modulates the hidden-state update when an observation event arrives.
In Latent ODE, it modulates the recognition process that estimates the initial latent distribution.
The backbone generates a new reconstruction and updates the cache for the next internal prediction.
During training, PCD retains the original reconstruction objective and introduces an additional energy term that constrains weighted prediction errors.
PCD combines trajectory generation, observation comparison, error weighting, and state correction while preserving the original continuous dynamic propagation.
The following sections define the problem formulation, introduce the continuous-time backbones, describe predictive coding signal construction, and present PCD integration and optimization.

\subsection{Problem Definition and Continuous-Time Backbone}
\label{subsec:problem-backbone}

Let the observation set of an irregular multivariate time series be
\begin{equation}
    \mathcal{O}
    = \{(t_i, \mathbf{x}_i, \mathbf{m}_i)\}_{i=1}^{T},
    \label{eq:observation-set}
\end{equation}
where $t_i \in \mathbb{R}$ denotes the $i$-th observation time,
$\mathbf{x}_i \in \mathbb{R}^{D}$ denotes a $D$-dimensional observation vector,
and $\mathbf{m}_i \in \{0,1\}^{D}$ denotes the observation mask.
If $m_{i,d}=1$, variable $d$ is observed at time $t_i$;
if $m_{i,d}=0$, the corresponding entry is missing.
Given a target time set
\begin{equation}
    \mathcal{T}_q = \{\tau_j\}_{j=1}^{T_q},
    \label{eq:query-times}
\end{equation}
the model estimates the target state
$\mathbf{y}_j \in \mathbb{R}^{D}$ at any $\tau_j$:
\begin{equation}
    \hat{\mathbf{y}}_j
    = F_{\Theta}(\mathcal{O}, \tau_j).
    \label{eq:reconstruction-model}
\end{equation}
Here, $F_{\Theta}$ denotes the continuous-time reconstruction model
parameterized by $\Theta$, and $\hat{\mathbf{y}}_j$ denotes
the prediction at query time $\tau_j$.

The continuous-time backbone describes system dynamics through a latent state trajectory.
Let $\mathbf{z}(t) \in \mathbb{R}^{H}$ be the latent state at time $t$,
where $H$ is the latent dimension.
Its evolution is defined by a neural ordinary differential equation:
\begin{equation}
    \frac{\mathrm{d}\mathbf{z}(t)}{\mathrm{d}t}
    = f_{\theta}\bigl(\mathbf{z}(t)\bigr).
    \label{eq:latent-dynamics}
\end{equation}
Here, $f_{\theta}$ is the parameterized neural function
that defines the latent dynamics with parameters $\theta$.
Given an initial latent state $\mathbf{z}_0$,
an ODE solver produces a latent trajectory on the target time set:
\begin{equation}
    \mathbf{z}(\mathcal{T}_q)
    = \operatorname{ODESolve}
      (f_{\theta}, \mathbf{z}_0, \mathcal{T}_q),
    \label{eq:latent-trajectory}
\end{equation}
which is mapped to the observation space by a decoder $g_{\psi}$:
\begin{equation}
    \hat{\mathbf{Y}}
    = g_{\psi}\bigl(\mathbf{z}(\mathcal{T}_q)\bigr).
    \label{eq:trajectory-decoder}
\end{equation}
Here, $\hat{\mathbf{Y}}=\{\hat{\mathbf{y}}_j\}_{j=1}^{T_q}$,
and $g_{\psi}$ denotes the decoder parameterized by $\psi$.

In Latent ODE, the initial state is inferred from the observation
sequence by a recognition network:
\begin{align}
    (\boldsymbol{\mu}_0, \boldsymbol{\sigma}_0)
    &= \operatorname{Enc}_{\phi}(\mathbf{U}, \mathbf{t}),
    \label{eq:latent-encoder}\\
    \mathbf{z}_0
    &= \boldsymbol{\mu}_0
       + \boldsymbol{\sigma}_0 \odot \boldsymbol{\epsilon},
       \quad
       \boldsymbol{\epsilon}
       \sim \mathcal{N}(\mathbf{0}, \mathbf{I}).
    \label{eq:latent-sampling}
\end{align}
Here, $\operatorname{Enc}_{\phi}$ is an encoder with parameters $\phi$,
$\mathbf{U}$ denotes the observation input provided to the encoder,
$\mathbf{t}=(t_1,\ldots,t_T)$ is the observation time sequence,
and $\boldsymbol{\mu}_0$ and $\boldsymbol{\sigma}_0$ are the mean
and standard deviation of the initial latent-variable distribution,
respectively.

In ODE-RNN, the hidden state is continuously propagated by an ODE
between adjacent observations and recurrently updated when an
observation event arrives:
\begin{align}
    \tilde{\mathbf{h}}_i
    &= \operatorname{ODESolve}
       (f_{\theta}, \mathbf{h}_{i-1}, t_{i-1}, t_i),
    \label{eq:ode-rnn-propagation}\\
    \mathbf{h}_i
    &= \operatorname{GRU}_{\phi}
       (\tilde{\mathbf{h}}_i, \mathbf{U}_i).
    \label{eq:ode-rnn-update}
\end{align}
Here, $\mathbf{h}_i$ denotes the hidden state after the update at time $t_i$,
$\tilde{\mathbf{h}}_i$ denotes the state obtained by continuous propagation
from the previous hidden state, and $\mathbf{U}_i$ denotes the input
representation of the $i$-th observation event.

These backbone models provide continuous-time reconstruction for irregular time series.
In their standard formulations, the reconstructed trajectory is optimized through likelihood objectives.
Prediction errors between reconstructed states and observations are not incorporated into the current dynamic inference process as feedback.
Predictive coding refinement incorporates this error into the backbone input, allowing continuous-time state estimation to use observations and prediction errors.

\subsection{Construction of Predictive Coding Signals}
\label{subsec:pc-signals}

Predictive coding refinement uses the reconstruction trajectory from the previous round as the prediction source for the current iteration.
Let the reconstruction cache obtained at round $r-1$ be
\begin{equation}
    \mathbf{C}^{(r-1)}
    = \{\mathbf{c}_i^{(r-1)}\}_{i=1}^{T},
    \label{eq:pc-cache}
\end{equation}
where $\mathbf{c}_i^{(r-1)} \in \mathbb{R}^{D}$ denotes the reconstruction
stored by the model at observation time $t_i$ from iteration $r-1$,
and $r$ denotes the predictive coding iteration index.
The cache is detached through the stop-gradient operation,
preventing gradients from propagating through previous reconstruction iterations.

Given the cached trajectory, the PC predictor generates the prediction
for iteration $r$:
\begin{equation}
    \mathbf{p}_i^{(r)}
    = P_{\eta}\!\left(
        \left[
            \operatorname{sg}\!\left(\mathbf{c}_i^{(r-1)}\right),
            t_i
        \right]
      \right).
    \label{eq:pc-prediction}
\end{equation}
Here, $\mathbf{p}_i^{(r)} \in \mathbb{R}^{D}$ denotes the internal prediction
at time $t_i$ in iteration $r$,
$P_{\eta}$ denotes the predictive coding network parameterized by $\eta$,
$[\cdot,\cdot]$ denotes feature concatenation,
and $\operatorname{sg}(\cdot)$ denotes the stop-gradient operation.
This prediction is used to measure the consistency between the reconstructed
trajectory and observed data during predictive coding refinement.

When the cache is unavailable, the initial prediction is generated
using temporal linear interpolation.
For an interior observation point $1<i<T$,
\begin{align}
    \mathbf{p}_i^{(0)}
    &= \alpha_i \mathbf{x}_{i-1}
       + (1-\alpha_i)\mathbf{x}_{i+1},
    \label{eq:pc-initial-prediction}\\
    \alpha_i
    &= \frac{t_{i+1}-t_i}
            {\max(t_{i+1}-t_{i-1}, \epsilon_t)}.
    \label{eq:pc-interpolation-weight}
\end{align}
Here, $\epsilon_t=10^{-8}$ prevents numerical instability
caused by zero time intervals.
Boundary points are initialized with neighboring observations,
namely $\mathbf{p}_1^{(0)}=\mathbf{x}_2$
and $\mathbf{p}_T^{(0)}=\mathbf{x}_{T-1}$.
This initialization provides the first prediction required for computing
prediction errors before a reconstruction cache is available.

Given the observation and prediction, the masked prediction error is defined as
\begin{equation}
    \mathbf{e}_i^{(r)}
    = \mathbf{m}_i \odot
      \bigl(\mathbf{x}_i-\mathbf{p}_i^{(r)}\bigr).
    \label{eq:pc-masked-error}
\end{equation}
Here, $\mathbf{e}_i^{(r)} \in \mathbb{R}^{D}$ denotes the prediction error
at time $t_i$, and $\odot$ denotes element-wise multiplication.
The mask $\mathbf{m}_i$ restricts the error computation to observed variables,
preventing missing entries from contributing to the feedback signal.
This error records the deviation between predictions and observations
across variables and time points.

Different variables impose different constraints on the continuous dynamic state.
The implementation introduces learnable dimension weights
$\mathbf{w}=(w_1,\ldots,w_D)$ and keeps them on the probability simplex
during optimization, namely $w_d \geq 0$
and $\sum_{d=1}^{D} w_d=1$.
The effective weight is defined with a minimum value constraint:
\begin{equation}
    \bar{w}_d
    = \epsilon_w + \max(1-D\epsilon_w,0)\,w_d.
    \label{eq:pc-effective-weight}
\end{equation}
Here, $\epsilon_w$ is the minimum dimension weight,
and $\bar{w}_d$ is the effective feedback weight for variable $d$.
This constraint avoids assigning zero weights to variables with large
prediction errors during predictive coding energy optimization.

The prediction error provided to the backbone is defined as
\begin{equation}
    \tilde{\mathbf{e}}_i^{(r)}
    = \gamma_e
      \left(
          \mathbf{e}_i^{(r)}
          \odot \sqrt{D\bar{\mathbf{w}}}
      \right),
    \label{eq:pc-scaled-error}
\end{equation}
where $\tilde{\mathbf{e}}_i^{(r)}$ is the scaled prediction error,
$\gamma_e$ is the global scaling coefficient of the error channel,
and $\bar{\mathbf{w}}=(\bar{w}_1,\ldots,\bar{w}_D)$.
The square root is applied element-wise.
The factor $\sqrt{D\bar{\mathbf{w}}}$ scales variables according to their
weights and maintains the error magnitude at a scale comparable
to the observation input.
This weighting mechanism assigns larger feedback contributions to
variables with greater influence on dynamic reconstruction.

\subsection{Predictive Coding Refinement Integrated with Continuous-Time Backbones}
\label{subsec:pc-integration}

PCD incorporates prediction errors into the recognition process
of the continuous-time backbone.
For the $r$-th predictive coding iteration, the original observation
and the error signal are concatenated as
\begin{align}
    \mathbf{x}_i^{\mathrm{pc},(r)}
    &= \bigl[
           \mathbf{x}_i,
           \tilde{\mathbf{e}}_i^{(r)}
       \bigr],
    \label{eq:pc-augmented-observation}\\
    \mathbf{m}_i^{\mathrm{pc}}
    &= [\mathbf{m}_i,\mathbf{m}_i],
    \label{eq:pc-augmented-mask}
\end{align}
and the encoder input is formed as
\begin{equation}
    \mathbf{U}_i^{\mathrm{pc},(r)}
    = \bigl[
          \mathbf{x}_i^{\mathrm{pc},(r)},
          \mathbf{m}_i^{\mathrm{pc}}
      \bigr].
    \label{eq:pc-encoder-input}
\end{equation}
Here, $\mathbf{x}_i^{\mathrm{pc},(r)} \in \mathbb{R}^{2D}$
contains the observation and prediction error,
$\mathbf{m}_i^{\mathrm{pc}} \in \{0,1\}^{2D}$ is the corresponding mask,
and $\mathbf{U}_i^{\mathrm{pc},(r)}$ denotes the input provided
to the backbone at time $t_i$.
The first group of channels contains observations,
and the second group contains prediction errors with their directions
and magnitudes.
This design incorporates prediction errors into state inference during
trajectory generation and avoids modifying only the decoded trajectory.

In the ODE-RNN backbone, PCD changes the input of the event update:
\begin{align}
    \tilde{\mathbf{h}}_i^{(r)}
    &= \operatorname{ODESolve}
       \bigl(
           f_{\theta},
           \mathbf{h}_{i-1}^{(r)},
           t_{i-1},
           t_i
       \bigr),
    \label{eq:pc-ode-rnn-propagation}\\
    \mathbf{h}_i^{(r)}
    &= \operatorname{GRU}_{\phi}
       \bigl(
           \tilde{\mathbf{h}}_i^{(r)},
           \mathbf{U}_i^{\mathrm{pc},(r)}
       \bigr).
    \label{eq:pc-ode-rnn-update}
\end{align}
Here, $\mathbf{h}_i^{(r)}$ denotes the hidden state after the observation
update at iteration $r$.
Because $\mathbf{U}_i^{\mathrm{pc},(r)}$ contains
$\tilde{\mathbf{e}}_i^{(r)}$, the hidden-state update depends on
observed values and prediction errors.
The error signal is propagated through subsequent continuous evolution
and decoding within the backbone.

In the Latent ODE backbone, PCD modifies the recognition input
used to estimate the initial latent distribution:
\begin{equation}
    \bigl(
        \boldsymbol{\mu}_0^{(r)},
        \boldsymbol{\sigma}_0^{(r)}
    \bigr)
    = \operatorname{Enc}_{\phi}
      \bigl(\mathbf{U}^{\mathrm{pc},(r)},\mathbf{t}\bigr),
    \label{eq:pc-latent-encoder}
\end{equation}
\begin{equation}
    \begin{aligned}
        \mathbf{z}_0^{(r)}
        &= \boldsymbol{\mu}_0^{(r)}
           + \boldsymbol{\sigma}_0^{(r)}
             \odot \boldsymbol{\epsilon},\\
        \boldsymbol{\epsilon}
        &\sim \mathcal{N}(\mathbf{0},\mathbf{I}),
    \end{aligned}
    \label{eq:pc-latent-sampling}
\end{equation}
\begin{equation}
    \hat{\mathbf{Y}}^{(r)}
    = g_{\psi}\!\left(
        \operatorname{ODESolve}
        \bigl(f_{\theta},\mathbf{z}_0^{(r)},\mathcal{T}_q\bigr)
      \right).
    \label{eq:pc-latent-reconstruction}
\end{equation}
Here,
$\mathbf{U}^{\mathrm{pc},(r)}
=\{\mathbf{U}_i^{\mathrm{pc},(r)}\}_{i=1}^{T}$,
$\boldsymbol{\mu}_0^{(r)}$ and $\boldsymbol{\sigma}_0^{(r)}$
denote the mean and standard deviation of the initial latent distribution
after incorporating prediction errors,
and $\hat{\mathbf{Y}}^{(r)}$ is the reconstruction trajectory
obtained at round $r$.

The same predictive coding procedure is applied to both ODE-RNN and Latent ODE.
Prediction error is incorporated into dynamic state estimation
as part of the recognition input.
For ODE-RNN, it is incorporated during hidden-state updates
at observation events.
For Latent ODE, it is incorporated when estimating the initial latent distribution.
Both backbones follow the same predictive coding procedure:
\begin{equation}
    \mathbf{C}^{(r-1)}
    \rightarrow \mathbf{P}^{(r)}
    \rightarrow \mathbf{E}^{(r)}
    \rightarrow \mathbf{U}^{\mathrm{pc},(r)}
    \rightarrow \hat{\mathbf{Y}}^{(r)}.
    \label{eq:pc-refinement-pipeline}
\end{equation}
Here, $\mathbf{P}^{(r)}=\{\mathbf{p}_i^{(r)}\}_{i=1}^{T}$
and $\mathbf{E}^{(r)}=\{\mathbf{e}_i^{(r)}\}_{i=1}^{T}$.
This procedure uses reconstruction errors as additional inputs during inference,
allowing prediction errors to participate in state estimation.

\subsection{Optimization Objective and Iterative Reconstruction Cache}
\label{subsec:optimization-cache}

Model training retains the original objective of the continuous-time backbone.
For deterministic ODE-RNN, the base objective is the masked Gaussian
reconstruction loss:
\begin{equation}
    \mathcal{L}_0
    = -\frac{1}{S}\sum_{s=1}^{S}
      \log p_{\sigma}
      \bigl(\mathbf{Y}\mid\hat{\mathbf{Y}}^{(s)}\bigr),
    \label{eq:base-loss-ode-rnn}
\end{equation}
where $\mathbf{Y}=\{\mathbf{y}_j\}_{j=1}^{T_q}$ is the target trajectory,
$\hat{\mathbf{Y}}^{(s)}$ is the prediction of the $s$-th trajectory sample,
$S$ is the number of sampled trajectories,
and $p_{\sigma}$ denotes the Gaussian likelihood whose variance
is controlled by the observation noise parameter $\sigma$.

For Latent ODE, the objective includes the reconstruction likelihood
and KL regularization of the initial latent distribution:
\begin{equation}
    \begin{aligned}
        \mathcal{L}_0
        ={}& -\log\left(\sum_{s=1}^{S}\exp(\ell_s)\right)\\
           & + \beta D_{\mathrm{KL}}\!\left(
               q_{\phi}(\mathbf{z}_0\mid\mathbf{U}^{\mathrm{pc}})
               \,\middle\|\,
               p(\mathbf{z}_0)
             \right),
    \end{aligned}
    \label{eq:base-loss-latent-ode}
\end{equation}
where
$\ell_s=\log p_{\sigma}
(\mathbf{Y}\mid\hat{\mathbf{Y}}^{(s)})$,
$\beta$ is the KL coefficient,
$q_{\phi}(\mathbf{z}_0\mid\mathbf{U}^{\mathrm{pc}})$
is the posterior over the initial latent variable inferred by the encoder,
and $p(\mathbf{z}_0)$ is the prior distribution.

The predictive coding energy constrains the weighted difference
between reconstructed trajectories and observed targets.
The main branch uses reconstruction-error energy:
\begin{equation}
    E_{\mathrm{pc}}
    =
    \frac{
        \displaystyle
        \sum_{s=1}^{S}
        \sum_{j=1}^{T_q}
        \sum_{d=1}^{D}
        m_{j,d}^{q}\,\bar{w}_d
        \bigl(\hat{y}_{j,d}^{(s)}-y_{j,d}\bigr)^2
    }{
        \displaystyle
        \max\!\left(
            S\sum_{j=1}^{T_q}\sum_{d=1}^{D}m_{j,d}^{q},
            1
        \right)
    }.
    \label{eq:pc-energy}
\end{equation}
Here, $m_{j,d}^{q}$ is the observation mask for variable $d$
at target time $\tau_j$,
and $\bar{w}_d$ is the effective predictive coding dimension weight.
This energy term uses the same variable weights as the error channel,
maintaining consistent weighting between optimization and feedback computation.

The weight of the predictive coding energy is controlled
by a learnable nonnegative coefficient:
\begin{equation}
    \lambda_{\mathrm{pc}}
    = \operatorname{softplus}(a_{\mathrm{pc}}),
    \label{eq:pc-energy-coefficient}
\end{equation}
where $a_{\mathrm{pc}}$ is an unconstrained scalar parameter.
To avoid numerical scale imbalance between $E_{\mathrm{pc}}$
and the base loss, the implementation uses a detached scale:
\begin{equation}
    s_{\mathrm{pc}}
    =
    \min\!\left(
        \frac{
            \operatorname{sg}(\mathcal{L}_0)
        }{
            \max\!\bigl(
                \operatorname{sg}(E_{\mathrm{pc}}),
                10^{-8}
            \bigr)
        },
        10^{8}
    \right).
    \label{eq:pc-detached-scale}
\end{equation}
Here, $s_{\mathrm{pc}}$ is used only for scale matching,
and gradients are detached from $\mathcal{L}_0$ and $E_{\mathrm{pc}}$
during this operation.

The final optimization objective is
\begin{equation}
    \mathcal{L}
    = \mathcal{L}_0
      + r_{\mathrm{pc}}\lambda_{\mathrm{pc}}
        s_{\mathrm{pc}}E_{\mathrm{pc}},
    \label{eq:total-loss}
\end{equation}
where $r_{\mathrm{pc}}$ is a fixed ratio coefficient
for predictive coding energy.
This objective incorporates prediction errors into two parts
of the training process.
Prediction errors are provided as additional inputs during backbone inference
and are constrained by the energy term during optimization.

The main branch also includes an output residual correction module:
\begin{equation}
    \begin{aligned}
        \hat{\mathbf{Y}}^{(r,+)}
        ={}& \hat{\mathbf{Y}}^{(r)}\\
           & + \gamma_c R_{\xi}\!\left(
               \left[
                   \hat{\mathbf{Y}}^{(r)},
                   \mathbf{P}^{(r)},
                   \mathbf{E}^{(r)},
                   \boldsymbol{\tau}
               \right]
             \right),
    \end{aligned}
    \label{eq:output-residual-correction}
\end{equation}
\begin{equation}
    \gamma_c
    = c_c\,\operatorname{softplus}(a_c).
    \label{eq:output-correction-scale}
\end{equation}
Here, $R_{\xi}$ is a residual correction network with parameters $\xi$,
$\boldsymbol{\tau}$ is the time feature aligned with the prediction grid,
$c_c$ is a fixed scaling coefficient,
and $a_c$ is a learnable scalar parameter.
This branch adjusts local output deviations when the observation grid
and prediction grid are aligned.
The predictive coding process is primarily implemented through
backbone inference with $\mathbf{U}^{\mathrm{pc}}$ as input.

After each inference iteration, the model updates the cache
with the current reconstruction:
\begin{equation}
    \bar{\mathbf{Y}}^{(r)}
    =
    \operatorname{sg}\!\left(
        \frac{1}{S}\sum_{s=1}^{S}
        \hat{\mathbf{Y}}^{(r,+,s)}
    \right).
    \label{eq:mean-reconstruction}
\end{equation}
Here, $\bar{\mathbf{Y}}^{(r)}$ denotes the mean reconstruction trajectory
obtained from sampled predictions.
If the previous cache exists, the cache is updated
by an exponential moving average:
\begin{equation}
    \mathbf{C}^{(r)}
    = m_c\mathbf{C}^{(r-1)}
      + (1-m_c)\bar{\mathbf{Y}}^{(r)}.
    \label{eq:cache-ema-update}
\end{equation}
Here, $m_c \in [0,1)$ is the cache momentum.
If no cache exists, the model directly sets
$\mathbf{C}^{(r)}=\bar{\mathbf{Y}}^{(r)}$.
The updated $\mathbf{C}^{(r)}$ is used in the next round
to generate $\mathbf{P}^{(r+1)}$ for subsequent predictive coding refinement.
The cache and prediction error provide information for updating state
estimates in subsequent continuous-time reconstruction iterations.

\section{Experiments}
\label{sec:experiments}

PCD is evaluated on simulated dynamic systems, real irregular multivariate
time series, and regular-grid forecasting benchmarks.
The experiments evaluate error feedback under incomplete histories,
sparse trajectory recovery, and different domains and sampling patterns.
We integrate PCD into ODE-RNN and Latent ODE using the original
continuous-time backbones.
This design evaluates the contribution of prediction errors to state
recognition under the original model structures.
This section introduces the experimental setup, presents comparison results,
and analyzes individual components through ablation studies.

\subsection{Experimental Setup}
\label{subsec:experimental-setup}

The evaluation covers physical dynamics, clinical monitoring, human motion,
meteorological observations, and general forecasting.
Table~\ref{tab:datasets} summarizes the domains, observation types,
and evaluation tasks of the datasets.

The MuJoCo data are derived from Hopper dynamics in the DeepMind Control Suite
and are used for forecasting with missing histories and sparse trajectory
recovery~\cite{pcd48,pcd49}.
In the forecasting task, models observe the first 50 states
and predict the next 10 states.
Historical observations are randomly removed at ratios of
30\%, 50\%, and 70\% to evaluate performance with different observation densities.
In the trajectory recovery task, models reconstruct the full continuous sequence
from only 10\%, 20\%, 30\%, or 50\% of the observation points.

The real-world irregular benchmarks include PhysioNet 2012,
Human Activity, and the United States Historical Climatology Network
(USHCN)~\cite{pcd50,pcd51,pcd52}.
For Human Activity, models predict the next 1,000 milliseconds
from the first 3,000 milliseconds to evaluate short-term motion forecasting.
For USHCN, models predict meteorological states for the next month
from 24 months of historical records, representing long-term environmental variation.
PhysioNet 2012 is used for short-horizon clinical forecasting,
where the next three timestamps are predicted from 36 hours of sparse ICU measurements.

We additionally include the Darts benchmark for evaluation
under regular sampling~\cite{pcd53}.
Darts contains regular time-series forecasting datasets from different
application domains, temporal scales, and sampling frequencies.
The benchmark includes series with periodic patterns, long-term trends,
and different forecasting difficulties.

\begin{table*}[!tp]
    \caption{Datasets and Evaluation Tasks}
    \label{tab:datasets}
    \centering
    \footnotesize
    \renewcommand{\arraystretch}{1.15}
    \begin{tabularx}{\textwidth}{
        @{}
        >{\raggedright\arraybackslash}p{0.22\textwidth}
        >{\raggedright\arraybackslash}p{0.18\textwidth}
        >{\raggedright\arraybackslash}X
        >{\raggedright\arraybackslash}p{0.20\textwidth}
        @{}
    }
        \toprule
        Dataset & Domain & Observation Type & Task \\
        \midrule
        MuJoCo~\cite{pcd48,pcd49}
        & Physical simulation
        & Multivariate trajectories
        & Forecasting and interpolation \\

        PhysioNet 2012~\cite{pcd50}
        & Clinical monitoring
        & Irregular ICU measurements
        & Forecasting and interpolation \\

        USHCN~\cite{pcd52}
        & Climate monitoring
        & Irregular temperature observations
        & Forecasting \\

        Human Activity~\cite{pcd51}
        & Human sensing
        & Irregular sensor observations
        & Forecasting \\

        Darts~\cite{pcd53}
        & Time series benchmark
        & Multiple regular time series
        & Forecasting \\
        \bottomrule
    \end{tabularx}
\end{table*}

\subsection{Main Results}
\label{subsec:main-results}

PCD reduces forecasting errors of continuous-time models under incomplete
observations, indicating that prediction errors can provide additional
information for latent state refinement.
The forecasting experiments evaluate residual feedback under different
temporal dynamics and observation patterns.

As shown in Table~\ref{tab:mujoco-forecasting}, adding PCD reduces
the forecasting error of ODE-RNN under different levels of missing
historical observations.
In the regular setting, ODE-RNN + PCD reduces the MSE from 0.3280
to 0.0047, and it maintains an MSE of 0.0080 even when 70\%
of historical observations are removed.
The substantial improvement on MuJoCo is mainly attributed to
the deterministic and smooth nature of physical dynamics.
In this setting, reconstruction errors contain informative signals about
latent-state deviation rather than random observation noise.
Therefore, residual feedback can effectively guide state correction
during inference.
As observation sparsity increases, the benefit of PCD remains because
the model can exploit consistency between generated trajectories
and available observations.

Table~\ref{tab:irregular-forecasting} reports lower errors on several
real irregular forecasting benchmarks after adding PCD.
On USHCN, PCD improves Latent ODE performance and achieves lower errors
than the original backbone, while on Human Activity,
ODE-RNN + PCD obtains the lowest MSE among all compared methods.
The results suggest that residual feedback remains applicable across
different temporal scales, with reconstruction errors providing
additional information for latent state estimation.

Table~\ref{tab:physionet-forecasting} further evaluates PCD under
sparse clinical forecasting conditions.
On PhysioNet 2012, PCD variants achieve competitive MAE results,
while several methods obtain lower MSE values.
The gap between MSE and MAE suggests that PCD is effective in reducing
general prediction deviation, while large individual errors caused
by abrupt physiological changes remain difficult to handle.
The forecasting results show that PCD reduces errors for continuous-time
backbones under several observation settings.
The limited gain on PhysioNet suggests that residual feedback is less
beneficial when prediction errors mainly arise from irregular
physiological variations.

\begin{table*}[!tp]
    \caption{MuJoCo Forecasting with Increasingly Sparse Histories}
    \label{tab:mujoco-forecasting}
    \centering
    \footnotesize
    \setlength{\tabcolsep}{3pt}
    \renewcommand{\arraystretch}{1.10}
    \begin{tabular*}{\textwidth}{
        @{}l@{\extracolsep{\fill}}cccc@{}
    }
        \toprule
        Method
        & Regular
        & 30\% Dropped
        & 50\% Dropped
        & 70\% Dropped \\
        \midrule
        GRU-$\Delta t$~\cite{pcd55}
        & $0.223 \pm 0.020$
        & $0.198 \pm 0.036$
        & $0.193 \pm 0.015$
        & $0.196 \pm 0.028$ \\

        GRU-D~\cite{pcd01}
        & $0.578 \pm 0.042$
        & $0.608 \pm 0.032$
        & $0.587 \pm 0.039$
        & $0.579 \pm 0.052$ \\

        GRU-ODE~\cite{pcd05}
        & $0.856 \pm 0.016$
        & $0.857 \pm 0.015$
        & $0.852 \pm 0.015$
        & $0.861 \pm 0.015$ \\

        ODE-RNN~\cite{pcd04}
        & $0.328 \pm 0.225$
        & $0.274 \pm 0.213$
        & $0.237 \pm 0.110$
        & $0.267 \pm 0.217$ \\

        Neural CDE~\cite{pcd06}
        & $0.028 \pm 0.002$
        & $0.027 \pm 0.000$
        & $0.027 \pm 0.001$
        & $0.026 \pm 0.001$ \\

        ANCDE~\cite{pcd56}
        & $0.026 \pm 0.001$
        & $0.025 \pm 0.001$
        & $0.025 \pm 0.001$
        & $0.024 \pm 0.001$ \\

        Latent ODE~\cite{pcd04}
        & $0.029 \pm 0.011$
        & $0.056 \pm 0.001$
        & $0.055 \pm 0.004$
        & $0.058 \pm 0.003$ \\

        EXIT~\cite{pcd27}
        & $0.026 \pm 0.000$
        & $0.025 \pm 0.004$
        & $0.026 \pm 0.000$
        & $0.026 \pm 0.001$ \\

        LEAP~\cite{pcd28}
        & $0.022 \pm 0.002$
        & $0.022 \pm 0.001$
        & $0.022 \pm 0.002$
        & $0.022 \pm 0.001$ \\

        Augmented-ODE~\cite{pcd57}
        & $0.055 \pm 0.004$
        & $0.056 \pm 0.004$
        & $0.057 \pm 0.005$
        & $0.057 \pm 0.005$ \\

        ACE-NODE~\cite{pcd58}
        & $0.039 \pm 0.003$
        & $0.053 \pm 0.007$
        & $0.053 \pm 0.005$
        & $0.052 \pm 0.006$ \\

        Neural SDE~\cite{pcd25}
        & $0.028 \pm 0.004$
        & $0.029 \pm 0.001$
        & $0.029 \pm 0.001$
        & $0.027 \pm 0.000$ \\

        Neural LSDE~\cite{pcd29}
        & $0.013 \pm 0.000$
        & $0.014 \pm 0.001$
        & $0.014 \pm 0.000$
        & \pcdThird{0.013 \pm 0.001} \\

        Neural LNSDE~\cite{pcd29}
        & $0.012 \pm 0.001$
        & $0.014 \pm 0.001$
        & $0.014 \pm 0.001$
        & $0.014 \pm 0.000$ \\

        Neural GSDE~\cite{pcd29}
        & $0.013 \pm 0.001$
        & \pcdThird{0.013 \pm 0.001}
        & \pcdThird{0.013 \pm 0.000}
        & $0.014 \pm 0.000$ \\

        KRNO~\cite{pcd54}
        & \pcdSecond{0.007 \pm 0.002}
        & \pcdSecond{0.008 \pm 0.002}
        & \pcdSecond{0.0114 \pm 0.004}
        & \pcdSecond{0.0115 \pm 0.002} \\
        \midrule
        ODE-RNN + PCD (Ours)
        & \pcdBest{0.0047 \pm 0.0007}
        & \pcdBest{0.0051 \pm 0.0002}
        & \pcdBest{0.0055 \pm 0.0008}
        & \pcdBest{0.0080 \pm 0.0020} \\

        Latent ODE (RNN enc.) + PCD (Ours)
        & \pcdThird{0.0075 \pm 0.0008}
        & $0.0138 \pm 0.0005$
        & $0.0210 \pm 0.0052$
        & $0.0312 \pm 0.0124$ \\

        Latent ODE (ODE enc.) + PCD (Ours)
        & $0.0081 \pm 0.0026$
        & $0.0136 \pm 0.0028$
        & $0.0151 \pm 0.0018$
        & $0.0189 \pm 0.0063$ \\
        \bottomrule
    \end{tabular*}
    \par\smallskip
    \begin{minipage}{\textwidth}
        \footnotesize
        The task predicts the next 10 states from the first 50 states
        of Hopper trajectories, with 0\%, 30\%, 50\%, or 70\%
        of historical observations removed.
        Values are test MSE (mean $\pm$ standard deviation over
        three random seeds); lower is better.
        Baseline results are taken from Table~5 of~\cite{pcd54},
        following the MuJoCo setting in Stable Neural SDEs~\cite{pcd29}.
        PCD results use the same setting.
        The best, second-best, and third-best means in each column
        are marked in bold, underlined, and italicized, respectively.
    \end{minipage}
\end{table*}

\begin{table*}[!tp]
    \caption{Irregular Forecasting on USHCN and Human Activity}
    \label{tab:irregular-forecasting}
    \centering
    \footnotesize
    \setlength{\tabcolsep}{3pt}
    \renewcommand{\arraystretch}{1.10}
    \begin{tabular*}{\textwidth}{
        @{}l@{\extracolsep{\fill}}cccc@{}
    }
        \toprule
        & \multicolumn{2}{c}{USHCN}
        & \multicolumn{2}{c}{Human Activity} \\
        \cmidrule(lr){2-3}
        \cmidrule(l){4-5}
        Method
        & MSE ($\times 10^{-1}$)
        & MAE ($\times 10^{-1}$)
        & MSE ($\times 10^{-3}$)
        & MAE ($\times 10^{-2}$) \\
        \midrule
        GRU-D~\cite{pcd01}
        & $5.54 \pm 0.38$
        & $3.40 \pm 0.28$
        & $2.94 \pm 0.05$
        & $3.51 \pm 0.06$ \\

        CRU~\cite{pcd09}
        & $6.09 \pm 0.17$
        & $3.54 \pm 0.18$
        & $6.97 \pm 0.78$
        & $6.30 \pm 0.47$ \\

        Crossformer~\cite{pcd59}
        & $5.25 \pm 0.04$
        & $3.27 \pm 0.09$
        & $4.29 \pm 0.20$
        & $4.89 \pm 0.17$ \\

        Latent ODE~\cite{pcd04}
        & $5.62 \pm 0.03$
        & $3.60 \pm 0.12$
        & $3.34 \pm 0.11$
        & $3.94 \pm 0.12$ \\

        DLinear~\cite{pcd38}
        & $6.21 \pm 0.00$
        & $3.88 \pm 0.02$
        & $4.03 \pm 0.01$
        & $4.21 \pm 0.01$ \\

        TimesNet~\cite{pcd60}
        & $5.58 \pm 0.05$
        & $3.60 \pm 0.04$
        & $3.12 \pm 0.01$
        & $3.56 \pm 0.02$ \\

        PatchTST~\cite{pcd39}
        & $5.75 \pm 0.01$
        & $3.57 \pm 0.02$
        & $4.29 \pm 0.14$
        & $4.80 \pm 0.09$ \\

        Graph WaveNet~\cite{pcd40}
        & $5.29 \pm 0.04$
        & $3.16 \pm 0.09$
        & $2.89 \pm 0.03$
        & $3.40 \pm 0.05$ \\

        MTCNN~\cite{pcd41}
        & $5.39 \pm 0.05$
        & $3.34 \pm 0.02$
        & $3.03 \pm 0.03$
        & $3.53 \pm 0.03$ \\

        StemGNN~\cite{pcd42}
        & $5.75 \pm 0.09$
        & $3.40 \pm 0.09$
        & $8.81 \pm 0.37$
        & $6.90 \pm 0.02$ \\

        CrossGNN~\cite{pcd43}
        & $5.66 \pm 0.04$
        & $3.53 \pm 0.05$
        & $3.03 \pm 0.10$
        & $3.48 \pm 0.08$ \\

        FourierGNN~\cite{pcd44}
        & $5.82 \pm 0.06$
        & $3.62 \pm 0.07$
        & $2.99 \pm 0.02$
        & $3.42 \pm 0.02$ \\

        SeFT~\cite{pcd31}
        & $5.80 \pm 0.19$
        & $3.70 \pm 0.11$
        & $12.20 \pm 0.17$
        & $8.43 \pm 0.05$ \\

        Raindrop~\cite{pcd33}
        & $5.78 \pm 0.22$
        & $3.67 \pm 0.17$
        & $14.92 \pm 0.14$
        & $9.45 \pm 0.05$ \\

        Warpformer~\cite{pcd61}
        & $5.25 \pm 0.05$
        & $3.23 \pm 0.05$
        & \pcdThird{2.79 \pm 0.04}
        & \pcdThird{3.39 \pm 0.03} \\

        mTAND~\cite{pcd32}
        & $5.33 \pm 0.05$
        & $3.26 \pm 0.10$
        & $3.22 \pm 0.07$
        & $3.81 \pm 0.07$ \\

        Neural Flow~\cite{pcd26}
        & $5.35 \pm 0.05$
        & $3.25 \pm 0.05$
        & $4.05 \pm 0.13$
        & $4.36 \pm 0.09$ \\

        T-PatchGNN~\cite{pcd34}
        & \pcdSecond{5.00 \pm 0.04}
        & $3.08 \pm 0.04$
        & \pcdSecond{2.66 \pm 0.03}
        & \pcdBest{3.15 \pm 0.02} \\

        KRNO~\cite{pcd54}
        & \pcdBest{4.95 \pm 0.08}
        & $3.06 \pm 0.08$
        & $2.85 \pm 0.03$
        & $3.46 \pm 0.02$ \\
        \midrule
        ODE-RNN + PCD (Ours)
        & $5.1527 \pm 0.0515$
        & \pcdSecond{2.9887 \pm 0.0261}
        & \pcdBest{2.616 \pm 0.050}
        & \pcdSecond{3.180 \pm 0.085} \\

        Latent ODE (RNN enc.) + PCD (Ours)
        & $5.0995 \pm 0.0317$
        & \pcdThird{3.0224 \pm 0.0523}
        & $4.429 \pm 0.175$
        & $4.733 \pm 0.124$ \\

        Latent ODE (ODE enc.) + PCD (Ours)
        & \pcdThird{5.0379 \pm 0.0334}
        & \pcdBest{2.9527 \pm 0.0318}
        & $3.203 \pm 0.127$
        & $3.789 \pm 0.153$ \\
        \bottomrule
    \end{tabular*}
    \par\smallskip
    \begin{minipage}{\textwidth}
        \footnotesize
        USHCN evaluates meteorological forecasting from historical
        climate records; Human Activity evaluates motion forecasting
        from irregular sensor observations.
        Values are mean $\pm$ standard deviation over three random seeds,
        with scale factors shown in the column headers; lower is better.
        Baseline results are taken from Table~4 of~\cite{pcd54}.
        PCD results use the same setting.
        The best, second-best, and third-best means in each metric column
        are marked in bold, underlined, and italicized, respectively.
    \end{minipage}
\end{table*}

\begin{table*}[!tp]
\begin{minipage}[t]{\columnwidth}
    \caption{Short-Horizon Clinical Forecasting on PhysioNet 2012}
    \label{tab:physionet-forecasting}
    \centering
    \footnotesize
    \setlength{\tabcolsep}{3pt}
    \renewcommand{\arraystretch}{1.10}
    \begin{tabularx}{\linewidth}{
        @{}>{\raggedright\arraybackslash}Xcc@{}
    }
        \toprule
        Method & MSE & MAE \\
        \midrule
        PrimeNet~\cite{pcd62}
        & $0.7953 \pm 0.0000$
        & $0.6859 \pm 0.0001$ \\

        NeuralFlows~\cite{pcd26}
        & $0.4056 \pm 0.0033$
        & $0.4466 \pm 0.0027$ \\

        CRU~\cite{pcd09}
        & $0.6179 \pm 0.0045$
        & $0.5778 \pm 0.0031$ \\

        mTAN~\cite{pcd32}
        & $0.3809 \pm 0.0043$
        & $0.4291 \pm 0.0035$ \\

        SeFT~\cite{pcd31}
        & $0.7721 \pm 0.0021$
        & $0.6760 \pm 0.0029$ \\

        GNeuralFlow~\cite{pcd63}
        & $0.8207 \pm 0.0310$
        & $0.6759 \pm 0.0100$ \\

        GRU-D~\cite{pcd01}
        & $0.3419 \pm 0.0029$
        & $0.3992 \pm 0.0011$ \\

        Raindrop~\cite{pcd33}
        & $0.3478 \pm 0.0019$
        & $0.4044 \pm 0.0020$ \\

        Warpformer~\cite{pcd61}
        & \pcdBest{0.3056 \pm 0.0011}
        & $0.3661 \pm 0.0016$ \\

        tPatchGNN~\cite{pcd34}
        & $0.3133 \pm 0.0053$
        & $0.3697 \pm 0.0049$ \\

        GraFITi~\cite{pcd64}
        & \pcdThird{0.3075 \pm 0.0015}
        & $0.3637 \pm 0.0036$ \\

        Hi-Patch~\cite{pcd36}
        & \pcdSecond{0.3071 \pm 0.0029}
        & $0.3675 \pm 0.0042$ \\

        KAFNet~\cite{pcd65}
        & $0.3164 \pm 0.0028$
        & $0.3715 \pm 0.0038$ \\

        APN~\cite{pcd66}
        & $0.3093 \pm 0.0011$
        & $0.3650 \pm 0.0026$ \\

        TFMixer~\cite{pcd67}
        & $0.3118 \pm 0.0023$
        & $0.3576 \pm 0.0019$ \\
        \midrule
        ODE-RNN + PCD (Ours)
        & $0.395433 \pm 0.001147$
        & \pcdBest{0.317790 \pm 0.000697} \\

        Latent ODE (RNN enc.) + PCD (Ours)
        & $0.397631 \pm 0.002353$
        & \pcdThird{0.320536 \pm 0.001729} \\

        Latent ODE (ODE enc.) + PCD (Ours)
        & $0.398001 \pm 0.002584$
        & \pcdSecond{0.318925 \pm 0.001646} \\
        \bottomrule
    \end{tabularx}
    \par\smallskip
    \begin{minipage}{\linewidth}
        \footnotesize
        The task predicts the next three timestamps from 36 hours
        of sparse ICU observations.
        Baseline results are taken from Table~1 of TFMixer~\cite{pcd67}.
        Values are mean $\pm$ standard deviation over three random seeds;
        lower is better.
        The best, second-best, and third-best means in each metric column
        are marked in bold, underlined, and italicized, respectively.
    \end{minipage}
\end{minipage}%
\hfill
\begin{minipage}[t]{\columnwidth}
    \caption{PhysioNet 2012 Interpolation with Irregular Observations}
    \label{tab:physionet-interpolation}
    \centering
    \footnotesize
    \setlength{\tabcolsep}{3pt}
    \renewcommand{\arraystretch}{1.10}
    \begin{tabularx}{\linewidth}{
        @{}>{\raggedright\arraybackslash}Xcc@{}
    }
        \toprule
        Method
        & MSE ($\times 10^{-3}$)
        & MAE ($\times 10^{-2}$) \\
        \midrule
        Trans-mean~\cite{pcd66}
        & $7.76 \pm 0.18$ & $4.92 \pm 0.07$ \\

        GRU-D~\cite{pcd01}
        & $6.18 \pm 0.23$ & $4.35 \pm 0.06$ \\

        SeFT~\cite{pcd31}
        & $9.46 \pm 0.12$ & $5.51 \pm 0.12$ \\

        Raindrop~\cite{pcd33}
        & $10.65 \pm 0.12$ & $5.80 \pm 0.07$ \\

        Warpformer~\cite{pcd61}
        & $6.37 \pm 0.34$ & $4.43 \pm 0.20$ \\

        mTAND~\cite{pcd32}
        & \pcdThird{5.65 \pm 0.08}
        & \pcdThird{4.18 \pm 0.07} \\

        Latent ODE~\cite{pcd04}
        & $6.84 \pm 0.34$ & $4.67 \pm 0.14$ \\

        Neural Flow~\cite{pcd26}
        & $6.77 \pm 0.06$ & $4.60 \pm 0.07$ \\

        CRU~\cite{pcd09}
        & $10.30 \pm 0.10$ & $5.47 \pm 0.05$ \\

        t-PatchGNN~\cite{pcd34}
        & \pcdSecond{4.75 \pm 0.03}
        & \pcdSecond{3.59 \pm 0.03} \\

        FPT~\cite{pcd47}
        & $12.24 \pm 0.07$ & $5.77 \pm 0.12$ \\

        Time-LLM~\cite{pcd45}
        & $12.43 \pm 0.08$ & $6.00 \pm 0.12$ \\

        ISTS-PLM-S~\cite{pcd46}
        & $7.74 \pm 0.35$ & $5.02 \pm 0.16$ \\

        ISTS-PLM-V~\cite{pcd46}
        & $7.12 \pm 0.30$ & $4.58 \pm 0.08$ \\

        ISTS-PLM~\cite{pcd46}
        & \pcdBest{4.55 \pm 0.08}
        & \pcdBest{3.37 \pm 0.02} \\
        \midrule
        ODE-RNN + PCD (Ours)
        & $6.81 \pm 0.04$ & $4.45 \pm 0.02$ \\

        Latent ODE (RNN enc.) + PCD (Ours)
        & $6.58 \pm 0.32$ & $4.44 \pm 0.14$ \\

        Latent ODE (ODE enc.) + PCD (Ours)
        & $7.00 \pm 0.11$ & $4.73 \pm 0.12$ \\
        \bottomrule
    \end{tabularx}
    \par\smallskip
    \begin{minipage}{\linewidth}
        \footnotesize
        The task reconstructs masked clinical measurements
        from sparse ICU observations.
        Baseline results are taken from Table~3 of~\cite{pcd46}.
        Values are mean $\pm$ standard deviation over three random seeds,
        with scale factors shown in the column headers;
        lower is better.
        The best, second-best, and third-best means in each column
        are marked in bold, underlined, and italicized, respectively.
    \end{minipage}
\end{minipage}
\end{table*}

PCD achieves lower reconstruction errors in sparse trajectory recovery
when observations provide sufficient constraints on the underlying dynamics.
Performance decreases when interpolation uncertainty becomes large.
The interpolation experiments evaluate prediction-error feedback
under limited observation settings.

Table~\ref{tab:mujoco-interpolation} reports lower reconstruction errors
on MuJoCo after incorporating PCD.
At the 50\% observation rate, Latent ODE (ODE encoder) + PCD
reduces the reported error from 0.285 to 0.0125
in the table's $10^{-2}$ MSE units.
This value is lower than the available reported results
of the compared methods in this setting.
Similar error reductions are observed at lower observation rates,
with Latent ODE (ODE encoder) + PCD obtaining values of
0.1867, 0.0792, and 0.0648 in the same units
under 10\%, 20\%, and 30\% observations, respectively.
The results suggest that repeated comparison between generated trajectories
and observations contributes to latent trajectory refinement.
Under extremely sparse conditions, several diffusion models obtain
lower reconstruction errors than PCD.
The two approaches rely on different modeling assumptions.
Diffusion models learn trajectory distributions from data,
while PCD updates latent states of continuous-time models
using observation feedback.

Table~\ref{tab:physionet-interpolation} evaluates PhysioNet interpolation,
which reconstructs missing clinical measurements from irregular ICU observations.
Clinical interpolation contains larger variations among individual trajectories
and less predictable state transitions than the simulated MuJoCo dynamics.
The PCD variants obtain higher errors than several interpolation methods
on this dataset, indicating that additional modeling components
may be required when observations provide limited constraints.
The results suggest that PCD is more suitable for continuous systems
with relatively stable dynamics.
Recovery tasks with strong stochastic variations may require
additional uncertainty modeling.

\begin{table*}[!tp]
    \caption{MuJoCo Trajectory Interpolation at Different Observation Rates}
    \label{tab:mujoco-interpolation}
    \centering
    \footnotesize
    \setlength{\tabcolsep}{3pt}
    \renewcommand{\arraystretch}{1.10}
    \begin{tabular*}{\textwidth}{
        @{}l@{\extracolsep{\fill}}cccc@{}
    }
        \toprule
        & \multicolumn{4}{c}{MSE ($\times 10^{-2}$)} \\
        \cmidrule(l){2-5}
        Method
        & 10\% Observed
        & 20\% Observed
        & 30\% Observed
        & 50\% Observed \\
        \midrule
        RNN $\Delta t$~\cite{pcd04}
        & $2.454$ & $1.714$ & $1.250$ & $0.785$ \\

        RNN GRU-D~\cite{pcd01}
        & $1.968$ & $1.421$ & $1.134$ & $0.748$ \\

        ODE-RNN~\cite{pcd04}
        & $1.647$ & $1.209$ & $0.986$ & $0.665$ \\

        ODE-RNN + STEER~\cite{pcd70}
        & $1.528$ & $1.090$ & $0.796$ & $0.603$ \\

        RNN-VAE~\cite{pcd04}
        & $6.514$ & $6.408$ & $6.305$ & $6.100$ \\

        Latent ODE (RNN enc.)~\cite{pcd04}
        & $2.477$ & $0.578$ & $2.768$ & $0.447$ \\

        Latent ODE (ODE enc.)~\cite{pcd04}
        & $0.360$ & $0.295$ & $0.300$ & $0.285$ \\

        Latent ODE (RNN enc.) + STEER~\cite{pcd70}
        & $2.310$ & $0.561$ & $2.534$ & $0.386$ \\

        Latent ODE (ODE enc.) + STEER~\cite{pcd70}
        & $0.230$ & $0.235$ & $0.246$ & $0.210$ \\

        Latent ODE + ACA~\cite{pcd71}
        & $0.310$ & $0.270$ & \textemdash & $0.260$ \\

        Latent ODE + MALI~\cite{pcd72}
        & $0.350$ & $0.270$ & \textemdash & $0.260$ \\

        NAOMI~\cite{pcd73}
        & $0.442$ & $0.232$ & $0.146$ & $0.093$ \\

        NRTSI~\cite{pcd74}
        & $0.406$ & $0.122$ & $0.063$
        & \pcdThird{0.026} \\

        Neural CDE~\cite{pcd06}
        & $1.352$ & $1.071$ & $0.835$ & \textemdash \\

        CSDI~\cite{pcd75}
        & $0.484$
        & \pcdThird{0.061}
        & \pcdBest{0.024}
        & \textemdash \\

        SSSDS4~\cite{pcd76}
        & $0.190$ & $0.100$ & $0.059$ & \textemdash \\

        Diffusion-TS~\cite{pcd68}
        & \pcdSecond{0.073}
        & \pcdSecond{0.043}
        & \pcdThird{0.037}
        & \textemdash \\

        SSD-TS~\cite{pcd69}
        & \pcdBest{0.065}
        & \pcdBest{0.0316}
        & \pcdSecond{0.027}
        & \textemdash \\
        \midrule
        ODE-RNN + PCD (Ours)
        & $1.0772$ & $0.9119$ & $0.7900$ & $0.5384$ \\

        Latent ODE (RNN enc.) + PCD (Ours)
        & $0.2060$ & $0.0920$ & $0.0840$
        & \pcdSecond{0.0170} \\

        Latent ODE (ODE enc.) + PCD (Ours)
        & \pcdThird{0.1867}
        & $0.0792$
        & $0.0648$
        & \pcdBest{0.0125} \\
        \bottomrule
    \end{tabular*}
    \par\smallskip
    \begin{minipage}{\textwidth}
        \footnotesize
        The task reconstructs full trajectories from 10\%, 20\%,
        30\%, or 50\% observed time points.
        Each entry represents MSE in units of $10^{-2}$;
        lower is better.
        RNN and ODE backbone results are taken from Table~3
        of~\cite{pcd04}.
        STEER, ACA, and MALI results are attributed to
        \cite{pcd70,pcd71,pcd72}, respectively.
        Other interpolation results are collected from Table~4
        of~\cite{pcd68} and Table~2 of~\cite{pcd69}.
        PCD results use the same observation rates.
        \textemdash{} indicates an unavailable reported result.
        The best, second-best, and third-best available results
        in each column are marked in bold, underlined,
        and italicized, respectively.
    \end{minipage}
\end{table*}

Additional experiments on regularly sampled Darts benchmarks are reported
in Section S1 and Table S1 of the supplementary material.

\subsection{Ablation Analysis}
\label{subsec:ablation-analysis}

\begin{figure}[!t]
    \centering
    \includegraphics[width=\columnwidth]
        {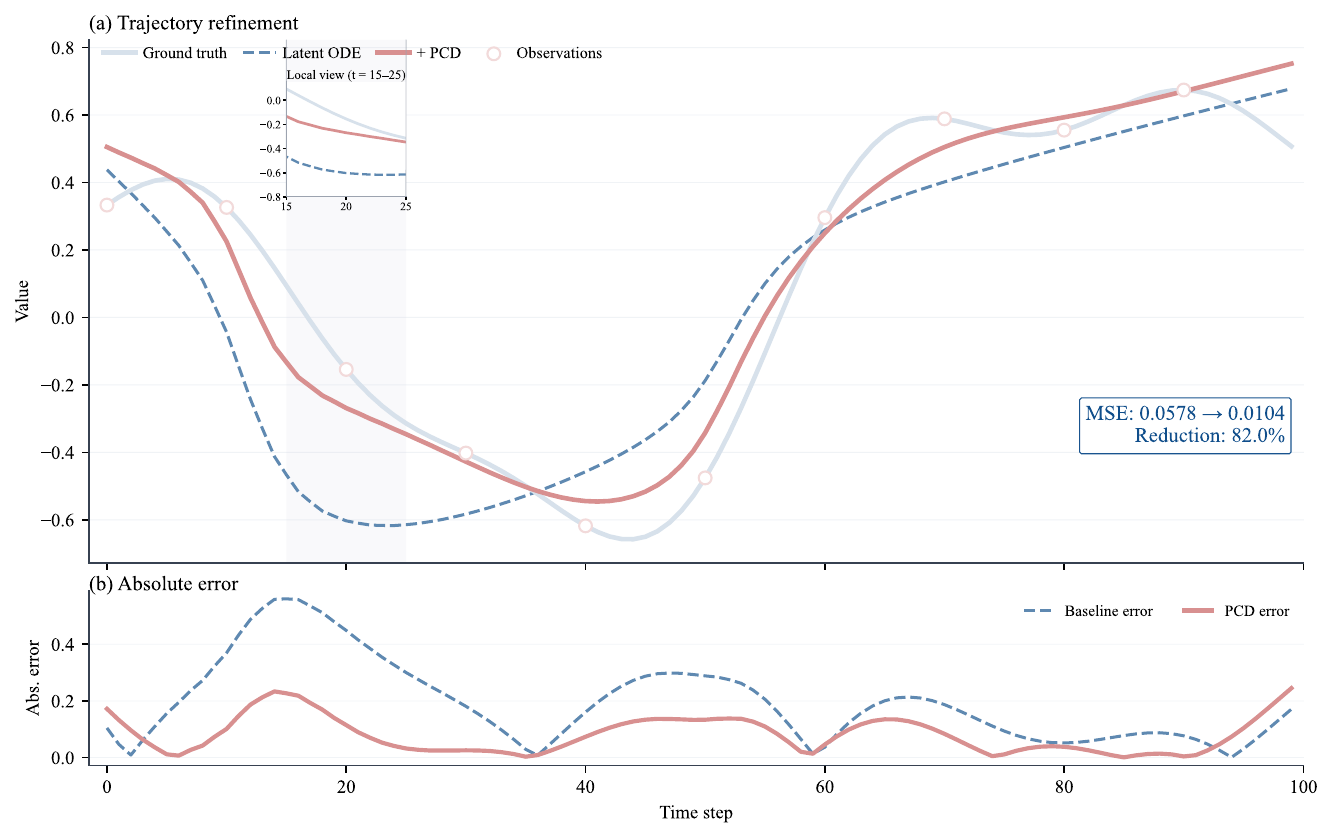}
    \caption{Visualization of predictive coding refinement
        on a synthetic one-dimensional trajectory.
        (a) Ground truth, Latent ODE baseline, and PCD-refined
        trajectories, with observations and a local zoom.
        (b) Absolute prediction errors of the baseline and PCD.
        PCD reduces the prediction deviation
        and follows the reference dynamics more closely.}
    \label{fig:synthetic-refinement}
\end{figure}

\begin{figure*}[!tp]
    \centering

    \subfloat[Latent ODE with an ODE encoder.]{%
        \includegraphics[width=0.49\textwidth]
            {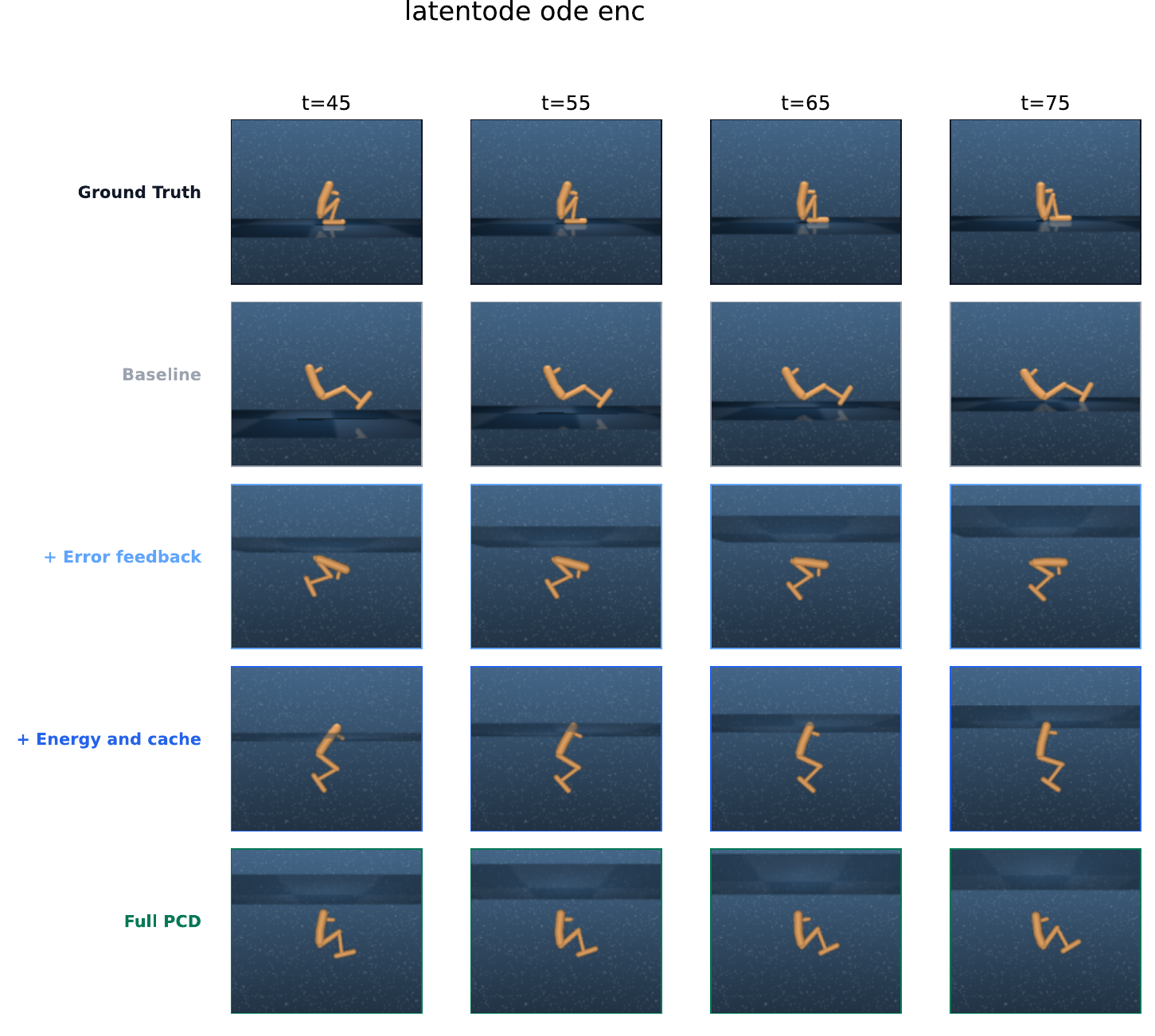}%
        \label{fig:sparse-latent-ode}%
    }%
    \hfill
    \subfloat[ODE-RNN.]{%
        \includegraphics[width=0.49\textwidth]
            {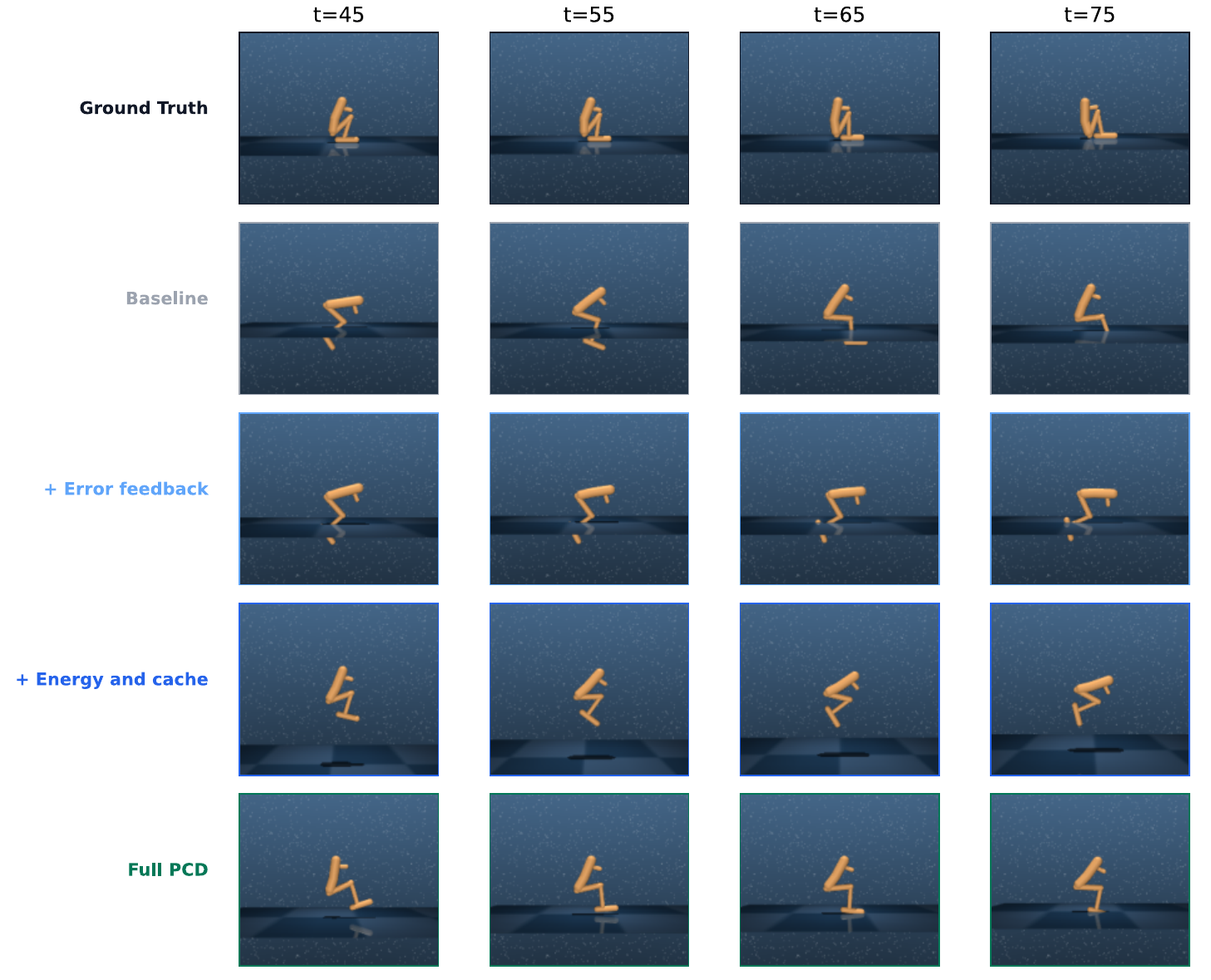}%
        \label{fig:sparse-ode-rnn}%
    }

    \caption{Trajectory refinement under sparse observations.
        Snapshots compare the ground truth, backbone predictions,
        and successive refinement configurations.
        Full PCD produces states closer to the ground truth
        than the corresponding backbone predictions
        in the illustrated examples.}
    \label{fig:sparse-refinement}
\end{figure*}

The performance of PCD is primarily associated with its residual feedback pathway.
Table~\ref{tab:component-ablation},
Fig.~\ref{fig:synthetic-refinement},
and Fig.~\ref{fig:sparse-refinement}
evaluate individual components and illustrate changes in trajectory estimation
after applying feedback.

Without the error feedback module, the forecasting MSE increases to
0.2210 and 0.2490 under 50\% and 70\% history dropping, respectively,
which is close to the original ODE-RNN performance.
Adding only the error feedback channel reduces these errors
to 0.0189 and 0.0295.
The ablation results reveal that the major performance gain originates
from incorporating residual information into the inference pathway
rather than merely adding additional regularization.
This indicates that prediction errors function as an active state estimation
signal instead of only a training objective.

The cache, energy constraint, variable weighting, and output correction modules
provide additional reductions in prediction error and adjust the contribution
of different variables during refinement.
The complete PCD model obtains MSE values of 0.0047, 0.0055, and 0.0080
under 0\%, 50\%, and 70\% history dropping, respectively.
Fig.~\ref{fig:synthetic-refinement} illustrates the reduction
of trajectory deviations after applying PCD.
Fig.~\ref{fig:sparse-refinement} shows the reconstructed trajectories
under sparse observations.
The results show that prediction errors are incorporated into latent state
estimation and are used during continuous-time inference.

\begin{table*}[!tp]
    \caption{Core Component Ablation on MuJoCo with the ODE-RNN Backbone}
    \label{tab:component-ablation}
    \centering
    \footnotesize
    \setlength{\tabcolsep}{3pt}
    \renewcommand{\arraystretch}{1.10}
    \begin{tabular*}{\textwidth}{
        @{}l@{\extracolsep{\fill}}ccccc@{}
    }
        \toprule
        & \multicolumn{3}{c}{Forecasting MSE}
        & \multicolumn{2}{c}{Reconstruction MSE ($\times 10^{-2}$)} \\
        \cmidrule(lr){2-4}
        \cmidrule(l){5-6}
        Variant
        & 0\% Dropped
        & 50\% Dropped
        & 70\% Dropped
        & 10\% Observed
        & 30\% Observed \\
        \midrule
        ODE-RNN Backbone
        & $0.3280 \pm 0.0220$
        & $0.2370 \pm 0.0180$
        & $0.2670 \pm 0.0250$
        & $1.647 \pm 0.061$
        & $0.986 \pm 0.038$ \\

        w/o Error Feedback
        & $0.2860 \pm 0.0190$
        & $0.2210 \pm 0.0170$
        & $0.2490 \pm 0.0220$
        & $1.582 \pm 0.056$
        & $0.951 \pm 0.034$ \\

        Error Feedback Only
        & $0.0118 \pm 0.0014$
        & $0.0189 \pm 0.0021$
        & $0.0295 \pm 0.0038$
        & $1.284 \pm 0.045$
        & $0.872 \pm 0.030$ \\

        w/o Cache
        & $0.0096 \pm 0.0011$
        & $0.0148 \pm 0.0017$
        & $0.0224 \pm 0.0030$
        & $1.226 \pm 0.042$
        & $0.849 \pm 0.027$ \\

        w/o Energy
        & $0.0074 \pm 0.0009$
        & $0.0102 \pm 0.0013$
        & $0.0157 \pm 0.0025$
        & $1.168 \pm 0.038$
        & $0.823 \pm 0.025$ \\

        w/o Variable Weight
        & $0.0063 \pm 0.0008$
        & $0.0081 \pm 0.0010$
        & $0.0119 \pm 0.0022$
        & $1.124 \pm 0.035$
        & $0.807 \pm 0.023$ \\

        w/o Output Correction
        & $0.0052 \pm 0.0007$
        & $0.0062 \pm 0.0009$
        & $0.0091 \pm 0.0021$
        & $1.095 \pm 0.033$
        & $0.798 \pm 0.022$ \\
        \midrule
        Full PCD
        & \pcdBest{0.0047 \pm 0.0007}
        & \pcdBest{0.0055 \pm 0.0008}
        & \pcdBest{0.0080 \pm 0.0020}
        & \pcdBest{1.077 \pm 0.031}
        & \pcdBest{0.790 \pm 0.021} \\
        \bottomrule
    \end{tabular*}
    \par\smallskip
    \begin{minipage}{\textwidth}
        \footnotesize
        Forecasting columns report test MSE under 0\%, 50\%,
        and 70\% history dropping.
        Reconstruction columns report test MSE in units of $10^{-2}$
        with 10\% and 30\% observed time points.
        Values are mean $\pm$ standard deviation over three random seeds;
        lower is better.
        The best result in each column is marked in bold.
        ``w/o'' denotes removal of the indicated component.
    \end{minipage}
\end{table*}

The following ablation experiments evaluate residual representations
and feedback configurations in PCD.
Table~\ref{tab:error-signal-ablation} compares different error signal formulations.
The signed residual provides the lowest forecasting error
among the evaluated variants.
Removing the error input increases the MSE to 0.2210.
Shuffled residuals reduce the error compared with no error input,
but the temporal and variable correspondence is lost.
Prediction-only input produces higher errors than residual feedback
because it lacks the deviation information between predictions and observations.
Squared and absolute errors yield higher errors because they remove
the directional information in residuals used for latent state adjustment.

\begin{table*}[!tp]
\begin{minipage}[t]{\columnwidth}
    \caption{Effect of the Prediction-Error Signal Form}
    \label{tab:error-signal-ablation}
    \centering
    \footnotesize
    \setlength{\tabcolsep}{3pt}
    \renewcommand{\arraystretch}{1.10}
    \begin{tabular*}{\columnwidth}{
        @{}l@{\extracolsep{\fill}}c@{}
    }
        \toprule
        Error-Signal Input
        & \shortstack{Forecast MSE\\50\% Dropped} \\
        \midrule
        No error input
        & $0.2210 \pm 0.0170$ \\

        Shuffled error
        & $0.1560 \pm 0.0130$ \\

        Prediction only
        & $0.0328 \pm 0.0031$ \\

        Squared error
        & $0.0117 \pm 0.0015$ \\

        Absolute error
        & $0.0094 \pm 0.0012$ \\

        Signed error
        & \pcdBest{0.0055 \pm 0.0008} \\
        \bottomrule
    \end{tabular*}
    \par\smallskip
    \begin{minipage}{\columnwidth}
        \footnotesize
        MuJoCo forecasting with 50\% history dropping.
        Shuffled error disrupts temporal and variable correspondence.
        Values are mean $\pm$ standard deviation over three random seeds;
        lower is better.
        The best result is marked in bold.
    \end{minipage}
\end{minipage}%
\hfill
\begin{minipage}[t]{\columnwidth}
    \caption{Effect of Prediction-Error Injection Location on MuJoCo}
    \label{tab:injection-location}
    \centering
    \footnotesize
    \setlength{\tabcolsep}{3pt}
    \renewcommand{\arraystretch}{1.12}
    \begin{tabularx}{\columnwidth}
        {@{}Xcc@{}}
        \toprule
        Injection Scheme
        & \shortstack{Forecast MSE\\50\% Dropped}
        & \shortstack{Reconstruction MSE\\($\times 10^{-2}$)\\20\% Observed} \\
        \midrule
        Backbone
        & $0.2370 \pm 0.0180$
        & $1.209 \pm 0.046$ \\

        Loss only
        & $0.0824 \pm 0.0068$
        & $1.078 \pm 0.039$ \\

        Output correction only
        & $0.0465 \pm 0.0042$
        & $0.991 \pm 0.034$ \\

        Encoder feedback only
        & $0.0062 \pm 0.0009$
        & $0.927 \pm 0.031$ \\

        Encoder feedback + output correction
        & \pcdBest{0.0055 \pm 0.0008}
        & \pcdBest{0.912 \pm 0.029} \\
        \bottomrule
    \end{tabularx}

    \par\smallskip
    \begin{minipage}{\columnwidth}
        \footnotesize
        Forecasting uses 50\% history dropping;
        reconstruction uses 20\% observed time points.
        Values are mean $\pm$ standard deviation
        over three random seeds; lower is better.
        The best result in each column is marked in bold.
    \end{minipage}
\end{minipage}
\end{table*}

Table~\ref{tab:injection-location} evaluates the location of residual feedback.
Introducing residual information into the recognition process yields lower
errors than applying output correction alone. This suggests that refining
the inferred state before dynamic propagation is more effective under the
evaluated settings. Additional cache-strategy comparisons and relative MSE
reductions are reported in Sections S2--S3 and Tables S2--S3 of the
supplementary material.

\section{Conclusion}
\label{sec:conclusion}

Real-world dynamic systems often produce sparse and irregular observations.
Conventional continuous-time models typically perform forward state
propagation and generate predictions without incorporating prediction
errors into trajectory estimation.
Under this formulation, initial estimation errors may accumulate
during long-horizon prediction.
This paper proposes Predictive Coding Dynamics (PCD)
to incorporate prediction-error feedback into continuous-time inference.
PCD stores previous reconstruction trajectories,
generates predictions at observation times,
constrains prediction errors through an energy term,
and provides these errors as additional inputs
to the continuous-time backbone.
This design incorporates prediction errors into state estimation
by comparing reconstructed trajectories with observations.

Experimental results show that error feedback reduces prediction errors
for the evaluated continuous-time architectures.
ODE-RNN and Latent ODE obtain lower errors on simulated dynamics,
clinical forecasting, and meteorological forecasting tasks
after incorporating PCD.
The error reductions are more apparent under sparse observation settings,
where forward propagation produces larger trajectory deviations.
PCD uses prediction errors from the model itself
as additional feedback signals during continuous state inference,
without introducing graph structures or additional attention modules.

The current predictive coding implementation has limitations
related to the interpretation of prediction errors.
The feedback signal directly uses residuals between predictions
and observations, which may combine true state drift
with sensor noise, anomalies, and random perturbations.
Such mixed errors may contain both persistent deviations
and transient noise, making error interpretation
during state updates difficult.
Although variable weighting adjusts the feedback contribution
of each variable, it does not fully capture complex error structures,
including dependencies among coupled variables
and temporal misalignment across different scales.
Performance is also constrained by the representational capacity
of the underlying continuous dynamics.
For highly nonstationary systems with long unobserved intervals,
prediction errors may provide limited information,
and the model relies mainly on backbone propagation.

Several directions remain for future investigation.
One direction is to develop predictive coding modules
that model uncertainty probabilistically.
Modeling errors as distributions may help separate observation noise
from state drift and improve performance under noisy observations.
Another direction is to model hierarchical and cross-variable
dependencies in feedback signals.
Multi-level predictive coding structures may support
different correction processes for slowly varying states
and short-term fluctuations.
Future studies may integrate prediction-error feedback
with state space models, diffusion generation models,
and pretrained time-series models
to evaluate its performance on larger sequence modeling tasks.

This work provides another perspective on irregular time-series modeling.
Recent studies have mainly focused on developing encoders
and continuous operators for mapping sparse inputs to future states.
Our results show that discrepancies between reconstructed trajectories
and observations provide additional information for state estimation.
Representing these discrepancies as feedback allows dynamic models
to update state estimates during inference.
Future forecasting systems may incorporate prediction-error feedback
to update internal states according to new observations.


\section*{Acknowledgment}

This work was supported by the National Science Fund
for Excellent Young Scholars (Overseas)
under Grant KZ37117501;
the National Natural Science Foundation of China
under Grant 62306024;
the National Cyber Security-National Science and Technology
Major Project under Grant 2025ZD1503602;
the New Generation Artificial Intelligence-National Science
and Technology Major Project
under Grants 2026ZD0128000 and 2026ZD0128003;
the Fundamental Research Funds for the Central Universities;
and the Beijing Advanced Innovation Center
for Future Blockchain and Privacy Computing.

\bibliographystyle{IEEEtran}
\bibliography{references-control,references}

\begin{IEEEbiography}[{\includegraphics[width=1in,height=1.25in,keepaspectratio]{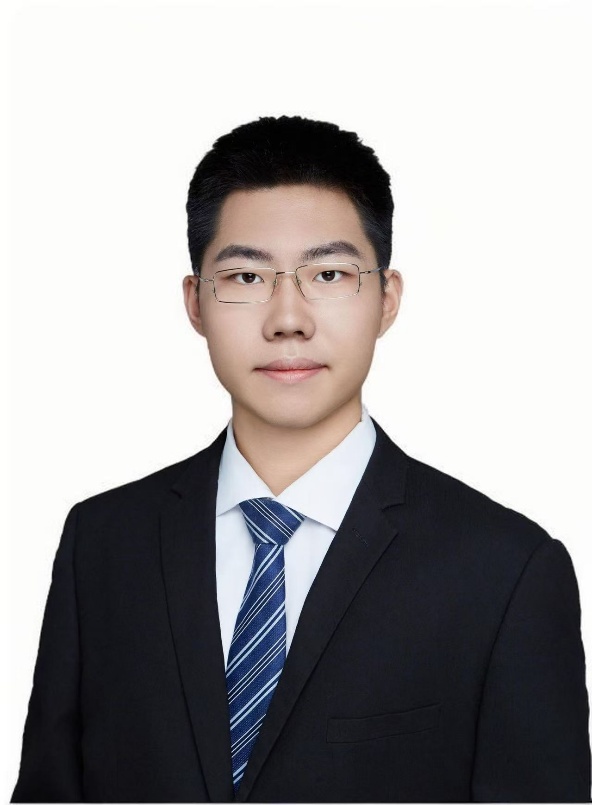}}]{Tianrui Gou}
received the B.S. degree from Beihang University, Beijing, China, in 2023.
He is currently pursuing the Ph.D. degree in artificial intelligence
with Beihang University, Beijing, China.
His research interests include brain-inspired computing and computational
neuroscience, with a focus on predictive coding and neural plasticity.
\end{IEEEbiography}

\begin{IEEEbiography}[{\includegraphics[width=1in,height=1.25in,keepaspectratio]{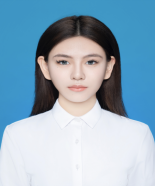}}]{Jiayangmei Xu}
received the B.A. degree from Beihang University in 2025,
where she is currently pursuing graduate studies in computational linguistics.
Her research focuses on machine learning and multimodal representation
learning, with an emphasis on view alignment and reinforcement learning
alignment for large language models.
\end{IEEEbiography}

\begin{IEEEbiography}[{\includegraphics[width=1in,height=1.25in,keepaspectratio]{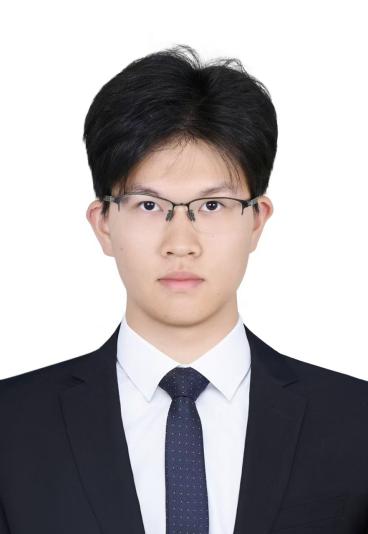}}]{Xingzhou Zhang}
is currently an undergraduate student majoring in information and computing
science with the School of Artificial Intelligence, Beihang University,
Beijing, China.
His research interests include deep learning, time series modeling,
and continuous-time dynamical systems.
\end{IEEEbiography}

\begin{IEEEbiography}[{\includegraphics[width=1in,height=1.25in,keepaspectratio]{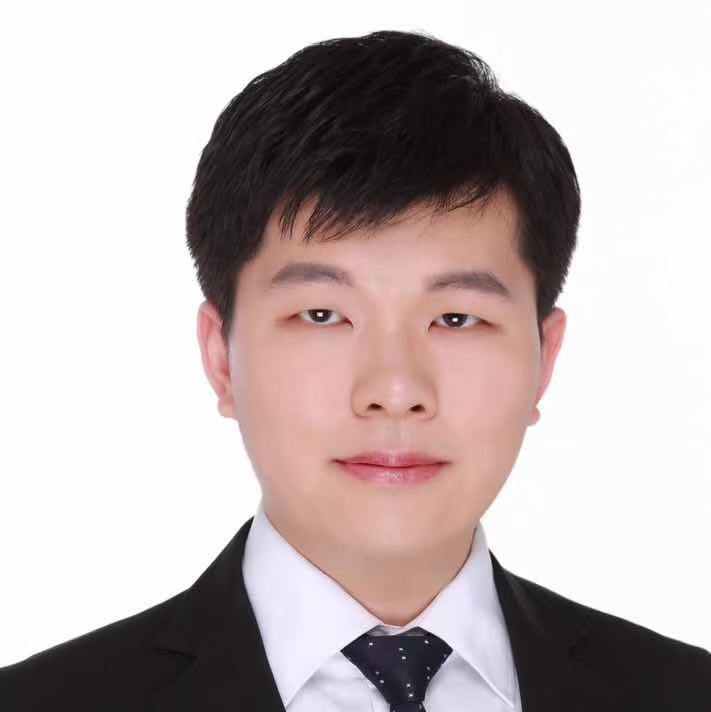}}]{Lei Sha}
is a professor and Ph.D. supervisor with the School of Artificial Intelligence,
Beihang University.
He received the Ph.D. degree in science from the Institute of Computational
Linguistics, School of Electronics Engineering and Computer Science,
Peking University, in 2018.
He subsequently joined Apple as a senior research scientist in the Siri NLP team.
From 2020 to 2022, he was a research associate with the Intelligent Systems
Group at the University of Oxford.
His research focuses on the development and maintenance of trustworthy large
language models and AI for Science.
He has published numerous papers in leading conferences and journals in the field.
He has served as an area chair for ACL, NeurIPS, and ICML and as a reviewer
for international journals.
He received the EMNLP 2020 Outstanding Reviewer Award and an ACL Outstanding
Paper Award.
\end{IEEEbiography}

\vfill
\end{document}